\documentclass[conference]{IEEEtran}
\IEEEoverridecommandlockouts

\usepackage{cite}
\usepackage{amsmath,amssymb,amsfonts}
\usepackage{graphicx}
\usepackage{textcomp}
\usepackage{subcaption}
\usepackage{xcolor}
\usepackage{soul}
\usepackage{color}
\usepackage{booktabs}
\usepackage{multirow}
\def\BibTeX{{\rm B\kern-.05em{\sc i\kern-.025em b}\kern-.08em
    T\kern-.1667em\lower.7ex\hbox{E}\kern-.125emX}}

\usepackage{flushend}

\begin{document}
\title{Dysphagia Risk Stratification in Head and Neck Cancer via Two-Stage PRO-Clinical Stacking}

\author{
\IEEEauthorblockN{
Siyuan Zhao\IEEEauthorrefmark{1},
Eric Ababio Anyimadu\IEEEauthorrefmark{2},
Zachary G. Brumm\IEEEauthorrefmark{3},
Yue Ma\IEEEauthorrefmark{3},\\
Clifton David Fuller\IEEEauthorrefmark{4},
Xinhua Zhang\IEEEauthorrefmark{1},
G. Elisabeta Marai\IEEEauthorrefmark{1},
Guadalupe Canahuate\IEEEauthorrefmark{2}
}

\IEEEauthorblockA{
\IEEEauthorrefmark{1}
\textit{Department of Computer Science, University of Illinois Chicago},
Chicago, IL, USA\\
\IEEEauthorrefmark{2}
\textit{Electrical and Computer Engineering, University of Iowa},
Iowa City, IA, USA\\
\IEEEauthorrefmark{3}
\textit{Department of Otolaryngology, University of California, San Francisco},
San Francisco, CA, USA\\
\IEEEauthorrefmark{4}
\textit{Department of Radiation Oncology, The University of Texas MD Anderson Cancer Center},
Houston, TX, USA
}

\IEEEauthorblockA{
\scriptsize
szhao69@uic.edu,
eric-anyimadu@uiowa.edu,
Zachary.Brumm@ucsf.edu,
Yue.Ma@ucsf.edu,\\
cdfuller@mdanderson.org,
zhangx@uic.edu,
gmarai@uic.edu,
guadalupe-canahuate@uiowa.edu
}
}

\maketitle

\begin{abstract}
Dysphagia is a debilitating late effect of head and neck cancer (HNC) treatment, yet timely identification of at-risk patients remains challenging in survivorship care. Definitive assessment relies on videofluoroscopic imaging, as captured by the Dynamic Imaging Grade of Swallowing Toxicity (CTCAE-DIGEST), which, while validated, requires specialized equipment, trained personnel, and significant patient burden, limiting its routine use in surveillance. Patient-reported outcomes (PROs), by contrast, are low-cost, scalable, and easily collected at any clinical encounter, making them an attractive alternative signal for identifying patients who may warrant further evaluation. However, a clear clinical framework for translating PRO responses into actionable interventions is still evolving. In particular, uncertainty remains regarding when a patient's self-reported symptom burden should prompt escalation of care.

This study addresses this gap by formulating a single-visit PRO-clinical prediction framework and introducing a clinically interpretable two-stage stacking model to predict swallowing impairment risk using PRO responses and structured clinical variables, without requiring videofluoroscopic imaging. The proposed framework quantifies the independent contributions of patient-reported symptoms and clinical factors within a unified and interpretable risk assessment model. Our findings demonstrate that individual MDADI responses contain predictive information beyond that captured by composite or global summary scores, while interpretability analyses reveal symptom patterns and clinical risk factors associated with swallowing impairment. Together, these results support the use of structured PRO-clinical integration as a practical, imaging-free approach for dysphagia risk stratification in HNC survivorship.
\end{abstract}

\vspace{12pt}

\begin{IEEEkeywords}
Patient-reported outcomes, Dynamic Imaging Grade of Swallowing Toxicity, head and neck cancer, XGBoost, ElasticNet, toxicity prediction
\end{IEEEkeywords}

\section{Introduction}
Dysphagia or swallowing impairment is among the most debilitating and persistent late effects of chemoradiation therapy for head and neck cancer (HNC), affecting an estimated 45--75\% of survivors and persisting or progressing years after completion of oncologic treatment~\cite{baudelet2019very,hutcheson2019two}. The Dynamic Imaging Grade of Swallowing Toxicity (CTCAE-DIGEST) integrates pharyngeal residue and airway invasion to provide a validated composite measure of swallowing impairment severity, and has been increasingly used as an imaging-based toxicity endpoint in HNC clinical trials and survivorship research~\cite{hutcheson2017dynamic,hutcheson2022refining}. Dysphagia after HNC treatment is associated with feeding tube dependence, aspiration pneumonia, and reduced quality of life, highlighting the clinical importance of identifying at-risk patients before functional decline becomes severe~\cite{hutcheson2012late,aylward2019rates,nguyen2005impact}.

Early identification of patients at risk for severe dysphagia before clinical signs emerge remains a critical unmet need in head and neck oncology. A predictive model capable of estimating swallowing impairment from routinely collected data could enable timely intervention, including referral for instrumental swallowing evaluation, prophylactic swallowing therapy, and dietary modification. Patient-reported outcomes (PROs), such as the MD Anderson Dysphagia Inventory (MDADI)~\cite{chen2001development}, are particularly attractive as predictive inputs because they provide a direct assessment of the patient's lived experience, capturing symptom burden and functional impairment from the patient's perspective in ways that are often not fully represented by clinician evaluations. Also, structured clinical variables, including tumor T/N staging, aspiration history, and treatment-related device use, may further complement self-reported symptoms by capturing objective prognostic information not accessible through PROs alone.

Prior machine learning studies in HNC have shown that PROs contain valuable predictive information for forecasting future symptom burden~\cite{anyimadu2024pro}. For example, Wang et al.\ demonstrated that LSTM-based modeling of longitudinal MDASI-HN responses~\cite{rosenthal2007measuring} improved prediction of late symptom severity relative to conventional approaches~\cite{wang2021predicting}. Despite this promise, the use of PROs for predicting clinically meaningful dysphagia risk remains relatively unexplored. Existing approaches have primarily focused on predicting continuous symptom severity scores rather than identifying patients at risk of dysphagia at levels warranting clinical intervention as defined by CTCAE-DIGEST. Furthermore, many of these methods rely on longitudinal PRO trajectories and incorporate limited clinical information, restricting their applicability in settings where only single-visit assessments are available.

Integrating PROs with clinical covariates presents a modeling challenge because these feature groups differ in scale, dimensionality, missingness, and statistical structure~\cite{anyimadu2025dual,anyimadu2025evaluating}. MDADI items are ordinal symptom measures that may contain nonlinear interactions, whereas clinical covariates are sparse mixed-type variables that require interpretable risk adjustment. Directly concatenating all features may impose a single modeling assumption on heterogeneous inputs, potentially diluting the PRO signal or overfitting sparse clinical variables. Moreover, while stacking offers a structured integration strategy, the meta-feature used by the second-stage model must be constructed carefully to avoid train-to-test distribution mismatch.

The main contributions of this study are threefold. First, we formulate a single-visit PRO-clinical prediction setting for swallowing impairment objectively defined using CTCAE-DIGEST Grade, targeting scenarios where longitudinal PRO questionnaires or complete treatment-planning data may be unavailable. 
{\color{black}Second, we propose a clinically motivated two-stage framework in which a PRO-derived risk score serves as the primary predictor and structured clinical covariates provide interpretable, incremental adjustment, enabling principled integration of heterogeneous PRO and clinical information.} Third, we evaluate the framework using 50-seed ablation experiments, comparative baselines, and interpretability analyses, demonstrating the predictive value of single-visit MDADI responses and characterizing the contributions of symptom items, composite scores, global scores, and clinical covariates.

\section{Related Work}
Prior work has established CTCAE-DIGEST as a validated framework for grading pharyngeal dysphagia in HNC, with the original scale combining swallowing safety and efficiency impairments into a CTCAE-compatible ordinal grade~\cite{hutcheson2017dynamic} and version~2 further refining these subscales and supporting the measure's ordinality and criterion validity~\cite{hutcheson2022refining}. Clinical studies have also reported associations between CTCAE-DIGEST-based impairment and PROs such as MDADI scores~\cite{pedersen2016swallowing,torkamani2015video,kendall2014quality,struder2023head,wishart2022association,liou2022evaluation,kirsh2019patient,tayebi2025mapping,andrade2017associaccao,van2023head,wentzel2024ditto,floricel2023roses,humbert2025externally,nipu2025lessons,nipu2025menergy,botto2025flavorcharter}, but these studies primarily focus on validation and association analysis rather than supervised prediction of CTCAE-DIGEST-defined swallowing dysfunction from item-level PRO responses.

Machine learning studies using PROs in HNC have more commonly focused on longitudinal symptom forecasting, including LSTM-based modeling of sequential MDASI-HN observations~\cite{wang2021predicting,rosenthal2007measuring} and sequential PRO-based frameworks for treatment-related toxicity prediction~\cite{anyimadu2024pro}. While these approaches demonstrate the predictive value of temporal PRO patterns, they often require repeated questionnaire measurements and typically predict patient-reported symptom severity or symptom-burden strata rather than imaging-derived CTCAE-DIGEST toxicity classes. Broader dysphagia and toxicity prediction models have relied on dosimetric, tumor-site, treatment-related, imaging, or organ-at-risk features~\cite{pu2025risk,goepfert2017predicting}, but these modalities may not be available during routine survivorship follow-up and are not specifically designed for single-visit PRO-based swallowing impairment classification.

Recent studies have explored compact PRO representations, showing that individual MDADI items can discriminate CTCAE-DIGEST-graded swallowing impairment~\cite{manduchi2025single}. However, single-item approaches may discard symptom-specific information distributed across the full questionnaire. Meanwhile, stacked generalization provides a methodological foundation for combining heterogeneous predictors via meta-features~\cite{wolpert1992stacked}, and probability-based meta-features have proven effective in classification stacking~\cite{ting1999issues}. Building on these directions, our work addresses the gap in single-visit swallowing impairment prediction by jointly modeling item-level MDADI responses and structured clinical covariates through a two-stage stacked framework that preserves nonlinear PRO information, supports interpretable clinical adjustment, and reduces dependence on longitudinal symptom histories.

\section{Data}
\label{sec:data}
Data were collected from patients enrolled in the University of California San Francisco (UCSF) Swallow Watch program, {\color{black}an automated dysphagia surveillance program for patients with head and neck cancer treated with radiotherapy}. The final cohort included 195 patients who contributed to 443 individual visit records. Eligible patients had a history of oropharyngeal or neck cancer and had received radiation therapy, chemoradiation, and/or surgical treatment. Patients were included if they completed the MDADI within 120 days of a videofluoroscopic swallow study (VFSS). Appropriate Institutional Review Board approval (IRB\#21-33649) was obtained, and the requirement for informed consent was waived due to the retrospective use of fully anonymized patient data.


Clinical covariates were selected under expert clinical guidance to capture routinely available factors associated with tumor burden, treatment exposure, airway or nutritional support, and swallowing-related complications. The final clinical feature vector included T/N staging codes, treatment-related variables, tracheostomy status, feeding-tube status, aspiration status and aspiration count, intravenous fluid use, and chemotherapy-related variables. Other clinically important variables, including HPV status and detailed radiation treatment-planning factors such as radiation dose, were considered but not included in the final model as they contained substantial missingness.

Swallowing-related PROs were collected using MDADI, a validated 20-item questionnaire assessing dysphagia-related quality of life in patients with HNC. Items cover global, emotional, functional, and physical domains and are scored on a 5-point Likert scale. Following standard scoring practice, the global item was excluded from composite score calculation; the remaining items were used to compute a composite score ranging from 20 to 100, with higher scores indicating better swallowing-related quality of life. MDADI responses were collected through a prospectively maintained REDCap database.

The outcome variable was the VFSS-derived CTCAE-DIGEST swallowing toxicity grade, which summarizes global pharyngeal dysphagia severity based on airway invasion and pharyngeal residue. CTCAE-DIGEST overall grades range from Grade~0, indicating no pharyngeal dysphagia, to Grade~4, indicating life-threatening swallowing toxicity requiring urgent intervention such as feeding-tube support. Specifically, Grade~1 represents mild dysphagia with asymptomatic or mild symptoms, Grade~2 represents moderate impairment that may limit normal daily activities, and Grade~3 represents severe, medically significant impairment affecting self-care. In this study, grades were dichotomized as Grade~0--1 versus Grade~2--4, with Grade~2--4 treated as the toxicity-positive group to identify clinically meaningful swallowing impairment.

\section{Methodology}

\subsection{Architecture}
We formulate swallowing impairment as a binary classification task based on the dichotomization defined in Section~\ref{sec:data}. Let 
$\mathcal{D} = \{(\mathbf{x}_i^{p}, \mathbf{x}_i^{c}, y_i)\}_{i=1}^{N}$ 
denote a dataset of $N$ visit-level records, where 
$\mathbf{x}_i^{p} \in \mathbb{R}^{20}$ is the MDADI PRO item vector, 
$\mathbf{x}_i^{c} \in \mathbb{R}^{10}$ is the clinical covariate vector, 
and $y_i \in \{0,1,2,3,4\}$ is the CTCAE-DIGEST grade. The binary outcome is defined as

\begin{equation}
  y_i^{\mathrm{bin}} = \mathbf{1}[y_i \geq 2],
  \label{eq:label}
\end{equation}

Each record belongs to a patient group $g_i \in \mathcal{G}$, where $\mathcal{G}$ denotes the set of patients and $g_i$ identifies the patient associated with visit $i$. The proposed two-stage stacking framework is trained and evaluated for this binary toxicity prediction task.

\begin{figure}[htbp]
\centering
\includegraphics[width=\columnwidth]{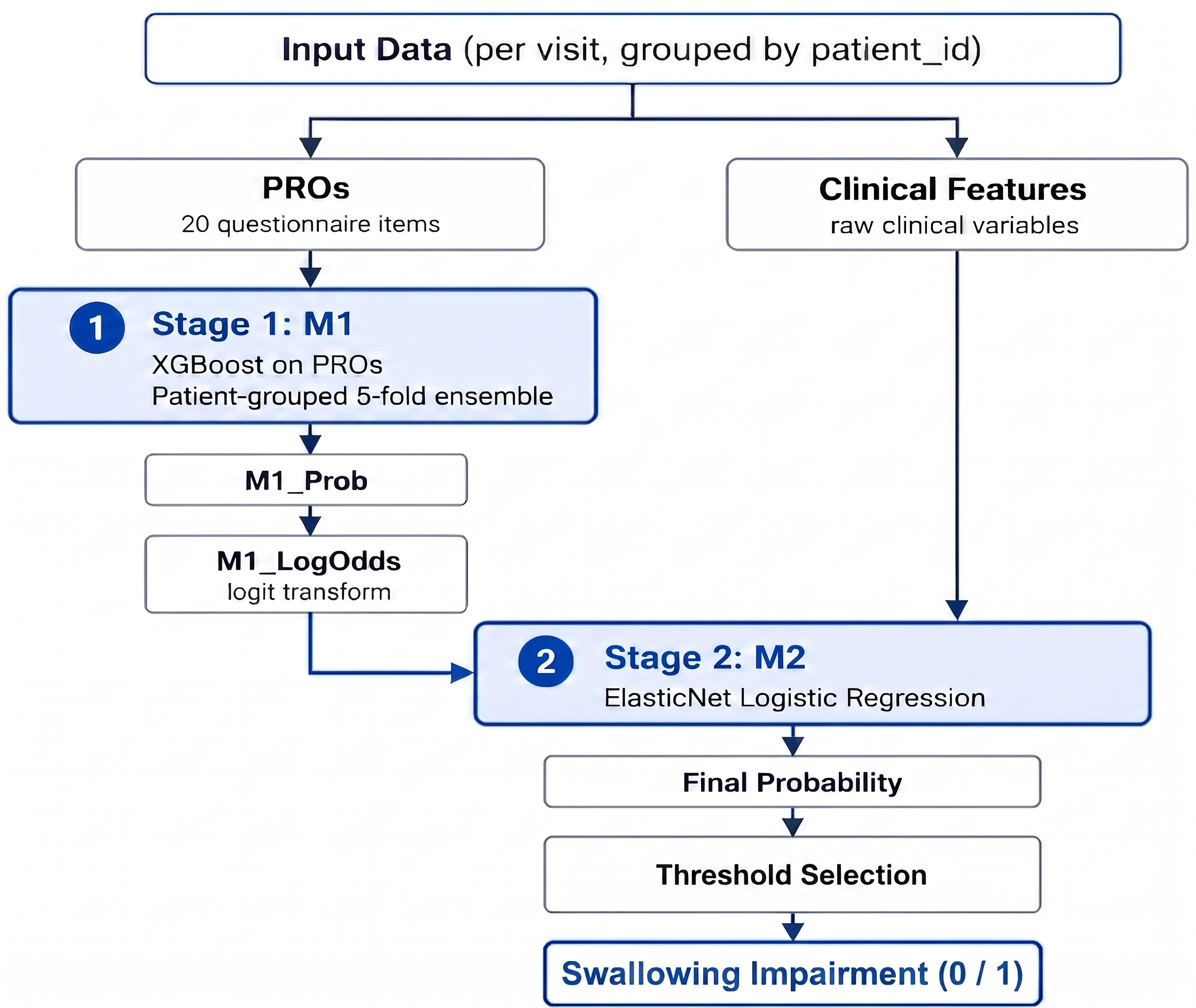}
\caption{Overview of the proposed two-stage framework for swallowing impairment prediction.}
\label{fig:architecture}
\end{figure}

The framework decomposes prediction into two sequential stages, as illustrated in Fig.~\ref{fig:architecture}. Stage~1 (M1) trains an XGBoost classifier~\cite{chen2016xgboost} exclusively on $\mathbf{x}^{p}$, producing a PRO-conditioned risk probability:

\begin{equation}
  \hat{p}_i^{(1)} = M_1(\mathbf{x}_i^{p}) \in [0,1].
  \label{eq:m1}
\end{equation}

Rather than retraining M1 on the full training set for test inference, we retain the five M1 models trained during 5-fold stratified group cross-validation, denoted as $\{M_1^{(k)}\}_{k=1}^K$ with $K{=}5$, and compute the test meta-feature as their ensemble average:

\begin{equation}
  \hat{p}_i^{(1)} = \frac{1}{K}\sum_{k=1}^{K} 
  M_1^{(k)}(\mathbf{x}_i^{p}), \quad i \in \mathcal{D}_\text{test}.
  \label{eq:ensemble}
\end{equation}

The raw probability is then mapped to the real line via the logit transform:

\begin{equation}
  \phi(\hat{p}_i^{(1)}) = \log\!\left(
  \frac{\hat{p}_i^{(1)}}{1 - \hat{p}_i^{(1)}}\right),
  \label{eq:logit}
\end{equation}
which aligns the meta-feature scale with the linear predictor of Stage~2 (M2). M2 integrates $\phi(\hat{p}_i^{(1)})$ with $\mathbf{x}_i^{c}$ via ElasticNet logistic regression~\cite{zou2005regularization}:

\begin{equation}
  \hat{p}_i^{(2)} = \sigma\!\left(\boldsymbol{\beta}^\top 
  \left[\phi(\hat{p}_i^{(1)}),\; \mathbf{x}_i^{c}\right]\right),
  \label{eq:m2}
\end{equation}
where $\sigma(\cdot)$ is the logistic function and $\boldsymbol{\beta}$ is estimated under the penalty:

\begin{equation}
  \mathcal{R}(\boldsymbol{\beta}) = 
  \frac{1-\alpha}{2}\|\boldsymbol{\beta}\|_2^2 \;+\; 
  \alpha\|\boldsymbol{\beta}\|_1, \quad \alpha = 0.5,\; C = 0.1.
  \label{eq:enet}
\end{equation}

The final binary prediction is obtained by thresholding $\hat{p}_i^{(2)}$ at $\tau^*$, selected from out-of-fold predictions by maximizing weighted F1 over $\tau \in [0.05, 0.95]$.

\subsection{Preprocessing}
The preprocessing pipeline was designed to address three properties of the dataset: repeated visits from the same patient, incomplete questionnaire and clinical entries, and heterogeneous clinical feature scales. All preprocessing steps were performed within the cross-validation workflow, with parameters estimated exclusively from training folds and applied to the corresponding validation folds, to avoid information leakage.

\subsubsection{Patient-Aware Data Splitting}
Because the dataset consists of visit-level records, multiple observations may belong to the same patient. Splitting data at the visit level could place records from the same patient in both training and validation sets, leading to overly optimistic performance estimates. To avoid this issue, we used 5-fold stratified group cross-validation with patient identifier as the grouping variable. This strategy assigns all visits from the same patient to a single fold while approximately preserving the class distribution across folds.

\subsubsection{Missing Value Imputation}
Missing values were handled according to the modeling requirements of each stage. Although XGBoost can handle missing values natively. PRO items in M1 were imputed using the training-fold mean to ensure a consistent feature representation across the five fold-specific models whose predictions are averaged to form the test meta-feature. For the ElasticNet model in M2, {\color{black}continuous variables were also imputed using the training-fold mean. Nominal categorical variables were first mode-imputed and then one-hot encoded. Categories not observed in the training fold were represented by an all-zero vector when they appeared in the corresponding validation fold.}

\subsubsection{Feature Standardization}
Clinical covariates include variables with different scales and measurement types. Since ElasticNet is sensitive to feature scale, continuous and ordinal clinical variables were standardized within each training-fold using $z$-score normalization. The same transformation was applied to the corresponding validation fold. PROs used by M1 were retained on their original scale because XGBoost, as a tree-based model, does not require feature standardization. After preprocessing, M1 received the original MDADI item values, while M2 received the PRO-derived risk representation together with imputed and standardized clinical covariates.

\subsubsection{Class Imbalance}
The positive class constitutes 25\% of training records, yielding an approximately 3:1 imbalance. M1 compensates via the \texttt{scale\_pos\_weight} hyperparameter, set to 3.0 to up-weight minority-class gradients during boosting. {\color{black}It was addressed only at M1, whereas M2 was trained with the standard logistic loss. This asymmetry is deliberate. Because M1's upweighting deliberately distorts its output probabilities toward the minority class, applying an additional class weight in M2 would compound this shift and degrade the calibration of the final predictions. Instead, M2 is refitted to the observed class distribution and thus serves as a recalibration step that maps the M1-derived log-odds and clinical covariates back onto empirical event frequencies, while class separation at deployment is governed by the selected threshold $\tau^\ast$ rather than by loss reweighting.}

\subsection{Experimental Setup}
We evaluated ten model variants, grouped by model family and input configuration, to quantify the contribution of each modeling component and to compare the proposed stacked design with simpler alternatives. These variants included clinical-only models, PRO-only models, direct PRO-clinical fusion models, composite-score baselines, and MDADI Global Score-based baselines using the "swallowing ability" item~\cite{md2019single} to assess whether a single symptom-focused item could provide performance comparable to the full PRO-based models.

\begin{itemize}
    \item \textbf{V1}: ElasticNet, clinical features only;
    \item \textbf{V2}: ElasticNet, all PRO items + clinical features;
    \item \textbf{V3}: ElasticNet, MDADI composite score + clinical features;
    \item \textbf{V4}: ElasticNet, swallowing ability + clinical features;
    \item \textbf{V5}: XGBoost, all PRO items only;
    \item \textbf{V6}: XGBoost, all PRO items + clinical features;
    \item \textbf{V7}: XGBoost, MDADI composite score + clinical features;
    \item \textbf{V8}: XGBoost, swallowing ability + clinical features;
    \item \textbf{V9}: XGBoost, all-item PRO-derived log-odds only;
    \item \textbf{V10}: Proposed Model, all-item PRO-derived log-odds + clinical features.
\end{itemize}

To obtain robust performance estimates, each variant was evaluated across 50 independent random seeds. For each seed, an outer patient-level split assigned approximately 70\% of patients to training and 30\% to testing; the same split was reused across all variants, enabling paired comparisons under identical partitions. Within the training set, an inner 5-fold stratified group cross-validation generated out-of-fold probabilities, and the decision threshold was selected by maximizing weighted F1 on these predictions. Each variant was then applied once to the held-out test set with the selected threshold. Performance was summarized as mean ± standard deviation across the 50 seeds, using AUC~\cite{hanley1982meaning}, weighted F1 (F1-W), balanced accuracy (Bal-Acc), and recall. Paired Wilcoxon signed-rank tests~\cite{woolson2007wilcoxon} compared V10 against each baseline across the 50 seed-level results, with significance at $p<0.05$. Effect sizes were quantified using paired Cohen's $d_z$.

\section{Results}

\subsection{Cohort Characteristics}
Table \ref{tab:clinical_treatment_characteristics} summarizes the demographic, clinical, and treatment characteristics of the study cohort. The cohort comprised 195 patients with a mean age of 64.54 years, of whom 74.9\% were male. Most patients presented with early-stage disease, with 55.9\% classified as T0--T2 and 53.8\% as N0--N1. Regarding treatment, 15.9\% of patients received chemoradiation, 11.8\% underwent surgery followed by radiation therapy, and 7.2\% received surgery combined with chemoradiation. The remaining patients underwent surgery alone or had treatment information categorized as unknown. With respect to chemotherapy, 12.3\% received concurrent definitive chemotherapy, 6.2\% received adjuvant chemotherapy, and 0.5\% received palliative chemotherapy. Clinical complications indicative of swallowing dysfunction were observed in a subset of patients. Feeding tube placement was required in 7.2\% of patients, while 4.6\% experienced one to three documented aspiration events. Additionally, 9.2\% required intravenous fluid support on one or more occasions. A total of 443 clinical visits were included in the analysis. Among these, 50.6\% were assigned a CTCAE-DIGEST grade of 0, indicating no clinically significant dysphagia, while the remaining visits were distributed across grades 1 through 4.

\begin{table}[htbp]
\centering
\caption{Summary of Patient Characteristics.}
\label{tab:clinical_treatment_characteristics}
\scriptsize
\setlength{\tabcolsep}{4pt}
\renewcommand{\arraystretch}{1.08}
\begin{tabular}{lll}
\toprule
\textbf{Variable} & \textbf{Category} & \textbf{Value} \\
\midrule

\multicolumn{2}{l}{\textbf{Age, mean (SD)}} & 64.54 (12.82)\\
\multicolumn{3}{l}{\textbf{Gender}} \\	
& Male	& 146 (74.9\%) \\
& Female & 49 (25.1\%) \\

\multicolumn{3}{l}{\textbf{Clinical T stage}} \\
& T0 & 13 (6.7\%) \\
& T1 & 49 (25.1\%) \\
& T2 & 47 (24.1\%) \\
& T3 & 21 (10.8\%) \\
& T4 & 24 (12.3\%) \\
& Unknown/missing & 41 (21.0\%) \\

\multicolumn{3}{l}{\textbf{Clinical N stage}} \\
& N0 & 42 (21.5\%) \\
& N1 & 63 (32.3\%) \\
& N2 & 39 (20.0\%) \\
& N3 & 10 (5.1\%) \\
& Unknown/missing & 41 (21.0\%) \\

\multicolumn{3}{l}{\textbf{Treatment type}} \\
& Chemoradiation alone & 31 (15.9\%) \\
& Radiation alone & 6 (3.1\%) \\
& Surgery alone & 4 (2.1\%) \\
& Surgery with CXRT & 14 (7.2\%) \\
& Surgery with XRT & 23 (11.8\%) \\
& Unknown/missing & 117 (60.0\%) \\

\multicolumn{3}{l}{\textbf{Chemotherapy types}} \\
& Adjuvant & 12 (6.2\%) \\
& Concurrent definitive & 24 (12.3\%) \\
& Palliative & 1 (0.5\%) \\
& No chemotherapy/Unknown & 158 (81.0\%) \\

\multicolumn{3}{l}{\textbf{Tracheostomy tube}} \\
& Yes & 1 (0.5\%) \\
& No & 194 (99.5\%) \\

\multicolumn{3}{l}{\textbf{Feeding tube}} \\
& Yes & 14 (7.2\%) \\
& No & 181 (92.8\%) \\

\multicolumn{3}{l}{\textbf{Aspiration incidence}} \\
& 0 & 186 (95.4\%) \\
& 1+ & 9 (4.6\%) \\

\multicolumn{3}{l}{\textbf{IV fluids received}} \\
& 0 & 85 (43.6\%) \\
& 1+ & 18 (9.2\%) \\
& Unknown/missing & 92 (47.2\%) \\

\addlinespace[0.5em]
\hline
\vspace{-0.8em}\\

\multicolumn{3}{l}{\textbf{CTCAE-DIGEST grades from VFSS studies}} \\

& Grade 0 & 224 (50.6\%) \\
& Grade 1 & 108 (24.4\%) \\
& Grade 2 & 58 (13.1\%) \\
& Grade 3 & 51 (11.5\%) \\
& Grade 4 & 2 (0.5\%) \\

\bottomrule
\end{tabular}
\end{table}

\subsection{Swallowing Impairment Prediction Performance}
The proposed model was evaluated for the binary swallowing impairment outcome derived from CTCAE-DIGEST grades as defined in Section~\ref{sec:data}. All model variants used the same feature definitions, preprocessing pipeline, model specifications, and evaluation protocol, ensuring that observed performance differences reflected differences in model design and input representation.

As shown in Table~\ref{tab:grade01_234_results}, the proposed stacked model achieved the strongest overall performance among the evaluated approaches. It obtained the highest values across all reported metrics, with the largest improvement observed in recall.

{\color{black}Against the weaker baselines (V1, V3, V4, V8), the proposed model showed large and significant gains across nearly all metrics ($d_z$ up to 3.7). Among the closest baselines, it was comparable on AUC to the strongest PRO-based variants (V5, V6, V9; $p>0.15$), suggesting that the clinical covariates primarily provided threshold-level adjustment rather than substantially changing the ranking. Its advantage was concentrated in balanced accuracy and recall, where it significantly exceeded the closest baselines (V2, V5, V6, V9; $p<0.01$, $d_z$ up to 0.83) and exceeded the composite-based V7 on AUC and balanced accuracy.}

The comparison across model variants also revealed substantial differences in performance according to the information source used for prediction. The clinical-only ElasticNet baseline achieved the lowest overall performance, and the MDADI composite-score and swallowing-ability baselines also demonstrated meaningful predictive performance. However, both reduced representations consistently performed worse than models that used the complete set of MDADI item responses.


\begin{table*}[t]
\caption{Model performance for swallowing impairment prediction. Values are reported as mean $\pm$ SD over 50 random seeds. {\bf V10: Proposed} corresponds to the proposed stacking model using M1: XGBoost over all-item PRO (Stage I) and M2: ElasticNet over M1 prediction + Clinical features (Stage II). Metrics include accuracy (Acc), area under the receiver operating characteristic curve (AUC), weighted F1 score (F1-W), balanced accuracy (Bal-Acc), and recall. Best results are highlighted in bold.}
\label{tab:grade01_234_results}
\centering
\scriptsize
\setlength{\tabcolsep}{2.5pt}
\renewcommand{\arraystretch}{1.08}
\resizebox{\textwidth}{!}{
\begin{tabular}{@{}lccccc@{}}
\toprule
\textbf{Model/Input} & \textbf{Acc} & \textbf{F1-W} & \textbf{AUC} & \textbf{Bal-Acc} & \textbf{Recall} \\
\midrule

V1: ElasticNet/Clinical-only
& $0.7200 \pm 0.0517$
& $0.6967 \pm 0.0556$
& $0.6930 \pm 0.0675$
& $0.5863 \pm 0.0578$
& $0.3102 \pm 0.1816$ \\

V2: ElasticNet/PRO+Clinical
& $0.8403 \pm 0.0327$
& $0.8367 \pm 0.0316$
& $0.8829 \pm 0.0370$
& $0.7640 \pm 0.0509$
& $0.6167 \pm 0.1149$ \\

V3: ElasticNet/Composite+Clinical
& $0.8273 \pm 0.0370$
& $0.8264 \pm 0.0369$
& $0.8779 \pm 0.0396$
& $0.7652 \pm 0.0610$
& $0.6439 \pm 0.1377$ \\

V4: ElasticNet/Swallowing+Clinical
& $0.7989 \pm 0.0353$
& $0.7972 \pm 0.0369$
& $0.8320 \pm 0.0468$
& $0.7225 \pm 0.0524$
& $0.5720 \pm 0.1114$ \\

V5: XGB/PRO-only
& $0.8310 \pm 0.0329$
& $0.8257 \pm 0.0333$
& $0.8845 \pm 0.0333$
& $0.7464 \pm 0.0576$
& $0.5817 \pm 0.1313$ \\

V6: XGB/PRO+Clinical
& $0.8386 \pm 0.0318$
& $0.8318 \pm 0.0350$
& $0.8820 \pm 0.0362$
& $0.7505 \pm 0.0572$
& $0.5783 \pm 0.1303$ \\

V7: XGB/Composite+Clinical
& $0.8350 \pm 0.0349$
& $0.8321 \pm 0.0369$
& $0.8703 \pm 0.0406$
& $0.7649 \pm 0.0620$
& $0.6285 \pm 0.1381$ \\

V8: XGB/Swallowing+Clinical
& $0.7664 \pm 0.0402$
& $0.7552 \pm 0.0537$
& $0.8160 \pm 0.0441$
& $0.6719 \pm 0.0795$
& $0.4817 \pm 0.2217$ \\

V9: XGB LogOdds/PRO-only
& $0.8294 \pm 0.0323$
& $0.8274 \pm 0.0303$
& $0.8845 \pm 0.0333$
& $0.7574 \pm 0.0510$
& $0.6205 \pm 0.1164$ \\

\textbf{V10: Proposed}
& $\mathbf{0.8463 \pm 0.0328}$
& $\mathbf{0.8448 \pm 0.0318}$
& $\mathbf{0.8850 \pm 0.0398}$
& $\mathbf{0.7825 \pm 0.0525}$
& $\mathbf{0.6616 \pm 0.1179}$ \\

\bottomrule
\end{tabular}
}
\end{table*}

{\color{black}
To confirm that model performance was not driven by a small subset of patients, we examined per-patient contribution and influence. Patients contributed a median of 2 visits (range 1–8), and the most frequently sampled 10\% of patients accounted for 26.0\% of all visits, indicating only a mild contribution imbalance. In the leave-one-patient-out jackknife analysis, removing any single patient changed the test AUC by at most 0.048 (std=0.017), and the most influential patients were those contributing only a single visit, indicating that influence did not increase with the number of visits a patient contributed. Further analysis indicates that repeated visits introduced a limited optimistic bias (modest reduction in performance: AUC 0.844 vs 0.885; F1-W 0.802 vs 0.845; balanced accuracy 0.756 vs 0.783) but did not drive the reported results (recall was essentially unchanged, 0.658 vs 0.662).}

{\color{black}We further compared the proposed model against a fixed composite-score screening rule ($\leq$ 60) currently used in clinical practice. The proposed model significantly outperformed this rule across all primary metrics: AUC (0.885 vs 0.872, p$<$0.001, d=0.50), weighted F1 (0.845 vs 0.813, p$<$0.001, d=1.33), balanced accuracy (0.783 vs 0.738, p$<$0.001, d=1.04), and recall (0.662 vs 0.589, p$<$0.001, d=0.74). The composite rule attained high specificity but substantially lower recall, reflecting that a single fixed cutoff misses a larger fraction of Grade~$\geq$2 cases. These results indicate that the proposed framework provides a measurable incremental benefit over the straightforward clinical decision rule, particularly in recall at the clinically relevant threshold.}

\subsection{Model Interpretability}
To examine how the proposed two-stage model generated its predictions, we conducted complementary interpretability analyses at each stage of the pipeline.

\subsubsection{Stage~1: PRO Feature Importance via SHAP}
\begin{figure}[htbp] 
\centering 
\includegraphics[width=\linewidth]{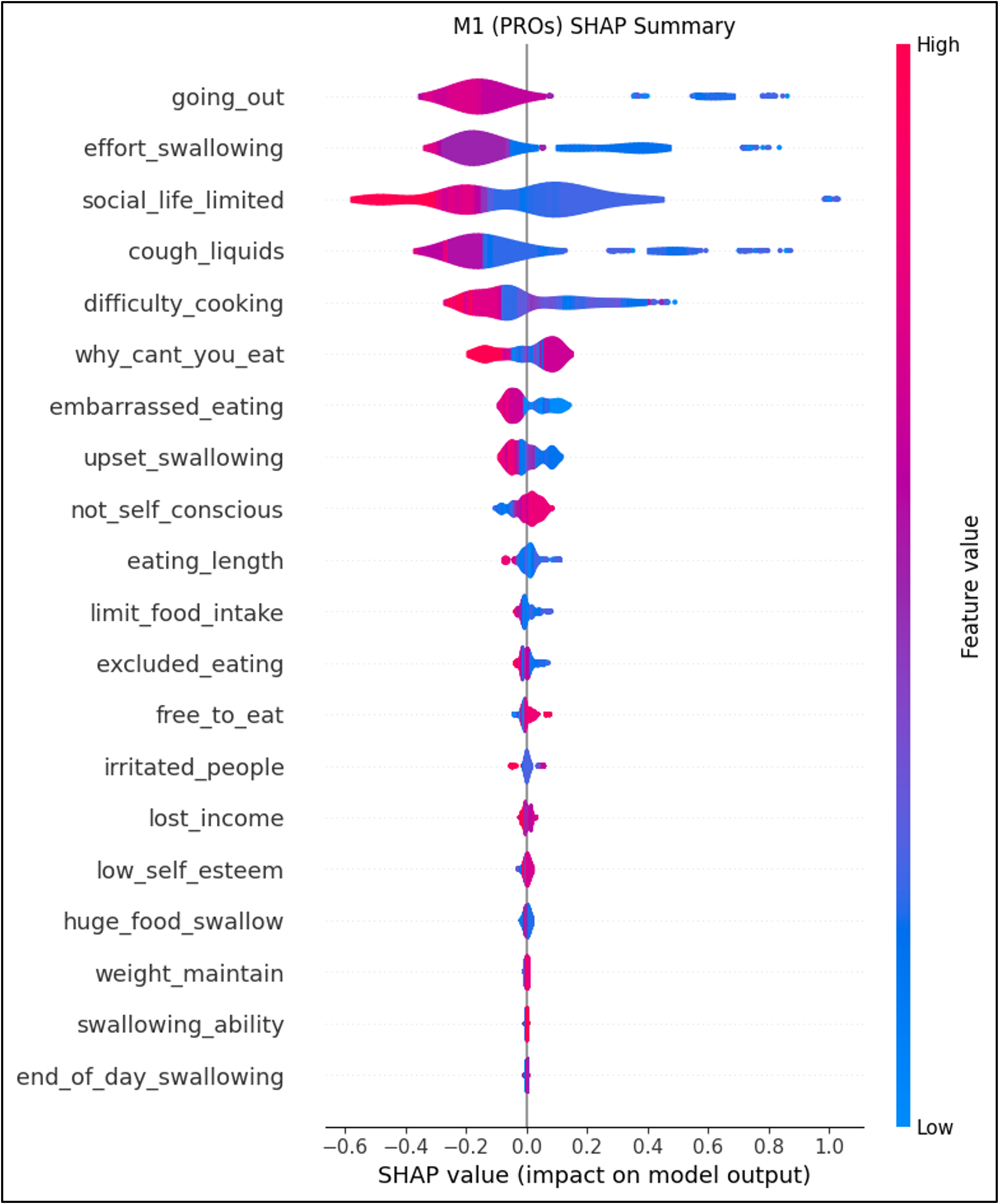} 
\caption{Stage~1 XGBoost SHAP feature importance plot.}
\label{fig:shap} 
\end{figure}

SHAP values~\cite{lundberg2017unified} were computed for the XGBoost-based M1 models and aggregated across the 5-fold stratified group cross-validation and all training visits. Figure~\ref{fig:shap} summarizes the resulting feature-attribution patterns. Lower MDADI item scores, shown in blue and corresponding to worse patient-reported swallowing function, were generally associated with positive SHAP values and therefore higher predicted toxicity risk. Higher scores, shown in red and corresponding to better function, were more often associated with negative SHAP values.

M1 relied most strongly on items reflecting functional limitations and social participation. \textit{Going\_out} ranked highest, followed by \textit{effort\_swallowing}, \textit{social\_life\_limited}, \textit{cough\_liquids}, and \textit{difficulty\_cooking}, whereas \textit{swallowing\_ability} ranked lower. This pattern suggests that clinically meaningful swallowing toxicity is more closely associated with the broader functional and social consequences of dysphagia than with global swallowing self-assessment alone, supporting the use of item-level MDADI responses as an informative predictor.

\subsubsection{Stage~2: ElasticNet Coefficient Stability Analysis}
\begin{figure}[htbp]
\centering
\includegraphics[width=\linewidth]{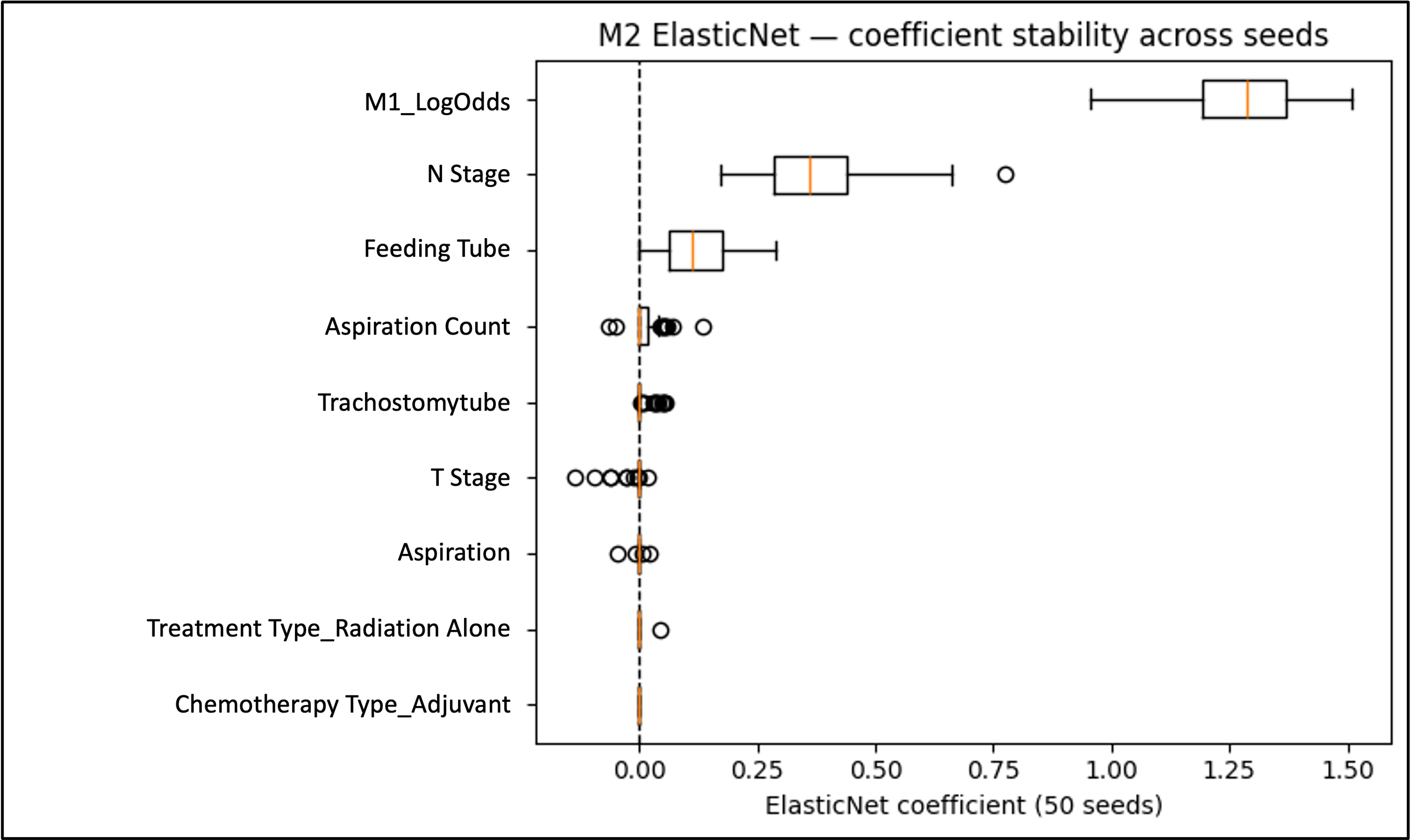}
\caption{Stage~2 ElasticNet coefficient distributions across 50 random seeds.}
\label{fig:coef}
\end{figure}

To interpret the contribution of the M2, ElasticNet coefficients were recorded across all 50 seeds. Each feature was summarized using three stability measures: selection frequency, defined as the proportion of seeds in which the feature received a nonzero coefficient; sign consistency, measuring whether the coefficient direction remained stable across seeds; and the distribution of nonzero coefficients. Figure~\ref{fig:coef} shows the resulting coefficient distributions.

\textit{M1\_LogOdds} was selected in all 50 seeds and showed the largest positive coefficient. This confirms that the PRO-derived risk estimate from M1 remained the dominant input to the final ElasticNet classifier. The result is clinically coherent, as swallowing impairment reflects clinically meaningful swallowing dysfunction that is likely to be perceived by patients and captured by MDADI responses.

Clinical variables played a secondary but still interpretable role. Among them, \textit{N Stage} was the most stable contributor and generally received positive coefficients, indicating that nodal disease status provided information not fully captured by PROs. This may reflect the relationship between higher nodal burden, more extensive regional treatment, and increased swallowing-related toxicity risk. \textit{Feeding Tube} also tended toward positive coefficients, consistent with its association with impaired swallowing function or greater nutritional support needs. By contrast, variables such as \textit{Treatment Type\_Radiation Alone} and \textit{Chemotherapy Type\_Adjuvant} were consistently removed by ElasticNet regularization, indicating limited independent predictive value after accounting for the PRO-derived risk score and other clinical covariates.

Overall, the coefficient stability analysis shows that M2 primarily relied on the PRO-derived risk score from M1, while selected clinical variables provided limited but meaningful complementary information. The most stable clinical contributors were related to disease burden and supportive care, whereas variables with inconsistent or near-zero coefficients should be interpreted cautiously. These results support the stacked design as an interpretable integration strategy: M1 captures the dominant symptom-derived signal, and M2 adjusts this estimate using clinically relevant covariates when they provide additional information.

\section{Discussion}
The central finding of this study is that low-cost, single-visit PRO-based swallowing symptoms can meaningfully identify patients at risk for clinically significant swallowing impairment, without the burden of videofluoroscopic imaging. While CTCAE-DIGEST remains the reference standard, it requires specialized equipment, trained personnel, and substantial patient effort. In contrast, the proposed framework relies on routinely collected PRO and clinical information and achieved strong predictive performance for identifying patients with moderate or greater swallowing toxicity. Importantly, the proposed model achieved the highest recall among all evaluated approaches. This is clinically relevant because Grade~2--4 CTCAE-DIGEST reflects swallowing impairment of sufficient severity to warrant closer surveillance, swallowing evaluation, or supportive intervention. Improved recall at this threshold may therefore facilitate earlier identification of patients at risk for clinically meaningful dysfunction.

Unlike structured clinical variables, PROs provide direct insight into patients' perceptions of symptom burden and functional limitations, capturing aspects of swallowing impairment that may not be fully reflected in clinician-assessed measures. These findings suggest that patient-reported symptoms contain substantial information regarding swallowing function and can serve as an effective surrogate source of risk information when objective swallowing assessments are unavailable. The proposed stacked framework, which combines PRO and clinical information, achieved the strongest overall performance, demonstrating that integrating patient-reported symptoms with clinical context offers a practical, imaging-free approach to dysphagia risk stratification.

The results also indicate that the manner in which PRO and clinical information are integrated is important. Direct feature-level fusion of PRO and clinical variables did not achieve the same performance as the proposed stacked framework, suggesting that simple concatenation may not fully exploit the complementary information contained within these heterogeneous data sources. In contrast, the proposed framework first learns complex symptom patterns from individual MDADI responses and then combines the resulting PRO-derived risk score with clinical covariates using an ElasticNet classifier. This design preserves the dominant predictive signal contained within the PRO responses while allowing clinical variables to provide interpretable risk adjustment. The resulting two-stage framework offers a compact and clinically interpretable strategy for integrating heterogeneous information sources.

Comparisons with composite-score and reduced-input baselines further contextualize the value of item-level PRO information. Although the MDADI composite score and the single swallowing-ability item both demonstrated meaningful predictive performance, neither consistently matched the performance of the full item-level model. This finding suggests that clinically meaningful swallowing impairment is influenced by multiple dimensions of patient-reported dysfunction, including symptom burden, functional limitation, and swallowing effort, which may not be fully captured by a single summary measure. Prior work has explored dimensionality reduction of the MDADI to identify compact symptom subscales, and the choice of which items to retain or aggregate may influence predictive performance. The present findings extend this line of inquiry by demonstrating that item-level responses provide a richer representation of swallowing burden than composite summaries and that individual symptom domains may contribute differently to dysphagia risk prediction.

The proposed method demonstrates that meaningful dysphagia risk stratification is achievable from a single clinical visit, requiring only a current PRO questionnaire and structured clinical record without longitudinal follow-up. This is practically significant, as repeated questionnaire collection across multiple visits increases burden for both patients and clinicians. Nevertheless, the results also highlight room for improvement, and it is reasonable to expect that incorporating longitudinal PRO trajectories, which capture symptom evolution over time, could further strengthen predictive performance. Leveraging repeated assessments to model functional decline therefore remains an important direction for future work.

This study has several limitations. First, the cohort was retrospective and from a single center, and external validation may be beneficial to ensure generalization of the findings. Second, although the 50-seed evaluation improves robustness, model performance may still be influenced by hyperparameter choices and threshold-selection criteria. Third, this study focused on binary CTCAE-DIGEST prediction as an initial step toward PRO-based dysphagia risk stratification. While this formulation provides a clinically practical framework for identifying patients at elevated risk of swallowing impairment, CTCAE-DIGEST encompasses multiple ordinal severity levels and integrates both safety and efficiency dimensions of swallowing function. As a foundational study, the proposed framework establishes a basis for future work exploring more granular prediction tasks, including ordinal severity modeling, multi-class classification, and component-specific assessment of swallowing toxicity. {\color{black}Future work includes further sensitivity, specificity, calibration metrics and decision-curve analysis, external validation, and prospective deployment.}

\section{Conclusion}
In conclusion, we introduced a clinically interpretable two-stage stacked learning framework for predicting the toxicity of swallowing impairment from single-visit MDADI responses and structured clinical features. The results suggest that single-visit PROs provide clinically relevant information beyond clinical variables alone for the prediction of swallowing impairment defined as CTCAE-DIGEST grade~2--4. These findings support the potential use of single-visit PRO-clinical modeling as a practical tool for dysphagia screening and follow-up decision support when longitudinal symptom data or imaging are unavailable.

\section*{Acknowledgment}
Our work is supported by NIH NCI R01CA258827, NIH UG3 TR004501, NSF CNS-2320261, and the UIC Institute for Health Data Science Research.

\bibliographystyle{IEEEtran}
\bibliography{references}

\end{document}